%% file: SingleImageGan.tex
\documentclass[10pt,twocolumn,letterpaper]{article}

\usepackage{iccv}
\usepackage{times}
\usepackage{epsfig}
\usepackage{graphicx}
\usepackage{amsmath}
\usepackage{amssymb}
\usepackage{caption}
\usepackage{multirow}

\usepackage[pagebackref=true,breaklinks=true,letterpaper=true,colorlinks,bookmarks=false]{hyperref}

\iccvfinalcopy 

\def\iccvPaperID{2245} 

\newcommand{\afterfig}{\vspace{-0.27cm}}
\ificcvfinal\fi
\begin{document}

\title{{{SinGAN: Learning a Generative Model from a Single Natural Image}}}
\author{Tamar Rott Shaham\\
Technion\\
\and
Tali Dekel\\
Google Research\\
\and 
Tomer Michaeli\\
Technion\\
}
\date{}

\twocolumn[{%
	\maketitle
	\vspace{-0.75cm}
	\renewcommand\twocolumn[1][]{#1}%
	\begin{center}
		\centering
		\includegraphics[width=1\textwidth]{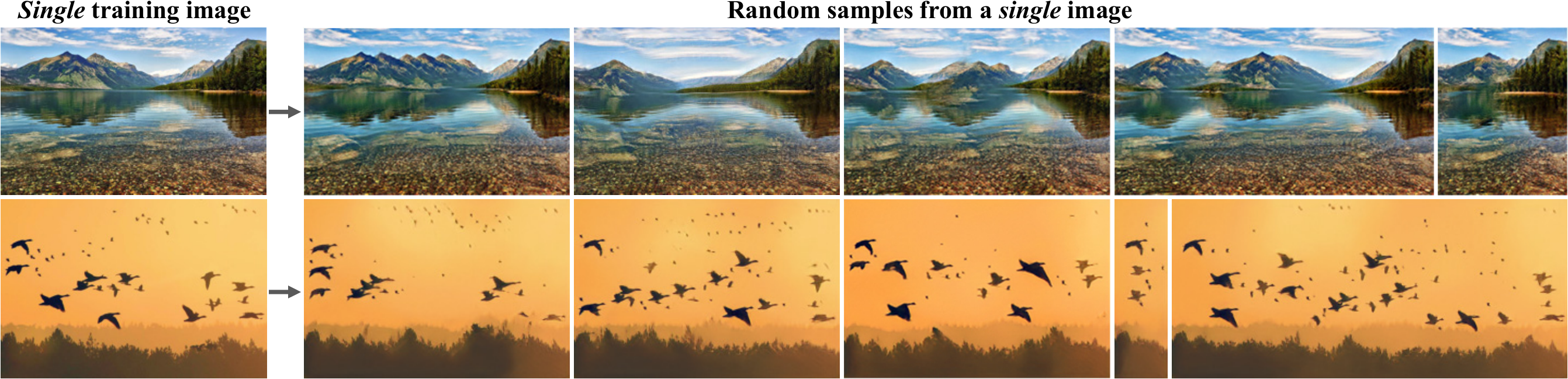}
		\captionof{figure}{\textbf{Image generation learned from a single training image.} We propose \emph{SinGAN}--a new unconditional generative model trained on a \emph{single natural image}. Our model learns the image's patch statistics across multiple scales, using a dedicated multi-scale adversarial training scheme; it can then be used to generate new realistic image samples that preserve the original patch distribution while creating new object configurations and structures. }
		\label{fig:teaser}
	\end{center}%
}]


\ificcvfinal\thispagestyle{empty}\fi

\input{abstract}

\input{intro_tali.tex}

\input{related}
\input{method}

\input{evaluation}

\input{applications}

\input{conclusions}

\vspace{0.15cm}
\noindent\textbf{Acknowledgements} Thanks to Idan Kligvasser for valuable insights. This research was
supported by the Israel Science Foundation (grant 852/17) and the Ollendorff foundation.
{\small
	\bibliographystyle{ieee_fullname}
	\bibliography{SingleImageGan}
}

\end{document}

%% file: abstract.tex
\begin{abstract}
We introduce \emph{SinGAN}, an unconditional  generative model 
that can be learned from 
a single natural image. 
Our model is trained to capture the internal distribution of patches within the image, and is then 
able to generate high quality, diverse samples 
that 
carry the same visual content 
as the image
. SinGAN contains a pyramid of fully convolutional GANs, each responsible for learning the patch distribution at a different scale of the image. This allows generating new samples of arbitrary size and aspect ratio, that have significant variability, yet maintain both the global structure and the fine textures of the training image. In contrast to previous single image GAN schemes, our approach is not limited to texture images, and is not conditional (\ie it generates samples from noise). User studies confirm that the generated samples are commonly confused to be real images. We illustrate the utility of SinGAN in a wide range of image manipulation tasks.
\end{abstract}

%% file: intro_tali.tex
\section{Introduction}


\begin{figure*}[t]
	\centering
	\includegraphics[width=1\textwidth]{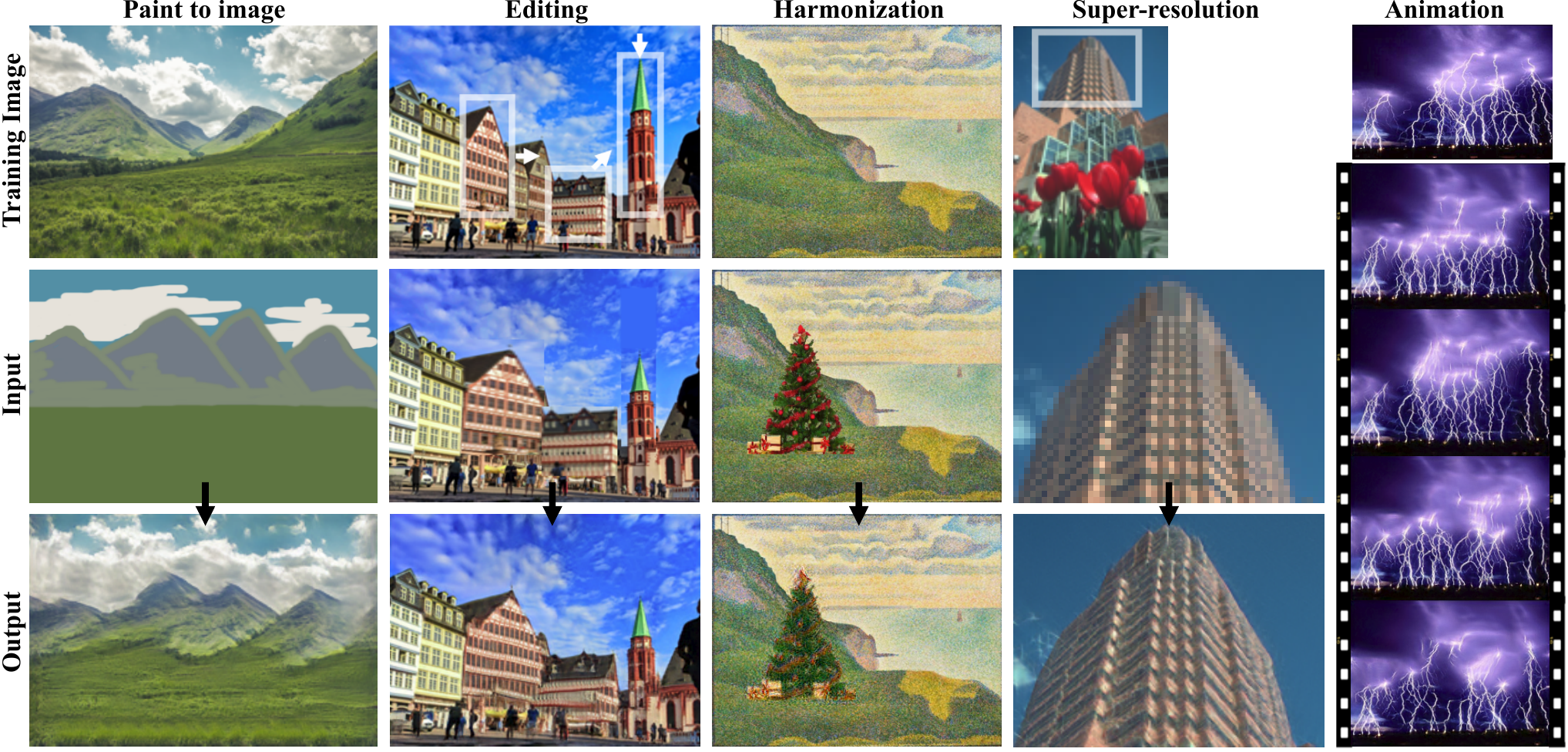}
	\caption{\textbf{Image manipulation.} SinGAN can be used in various image manipulation tasks, including: transforming a paint (clipart) into a realistic photo, rearranging and editing objects in the image, harmonizing a new object into an image, image super-resolution and creating an animation from a single input. In all these cases, our model observes only the training image (first row) and is trained in the same manner for all applications, with no architectural changes or further tuning (see Sec.~\ref{sec:applications}).}
	\label{fig:image_manipulation} \vspace{-0.35cm}
\end{figure*}


Generative Adversarial Nets (GANs) \cite{goodfellow2014generative} have made a dramatic leap in modeling high dimensional distributions of visual data. In particular, unconditional GANs have shown remarkable success in generating realistic, high quality samples when trained on class specific datasets (\eg, faces \cite{liu2015deep}, bedrooms\cite{silberman2012indoor}). However, capturing the distribution of highly diverse datasets with multiple object classes (\eg ImageNet \cite{imagenet_cvpr09}), is still considered a major challenge and often requires conditioning the generation on another input signal \cite{brock2018large} or training the model for a specific task (\eg super-resolution \cite{ledig2017photo}, inpainting \cite{pathak2016context}, retargeting \cite{shocher2018internal}).


Here, we take the use of GANs into a new realm -- \emph{unconditional} generation learned from a \emph{single natural image}. Specifically, we show that the internal statistics of patches within a single natural image typically carry enough information for learning a powerful generative model
. \emph{SinGAN}, our new single image generative model, allows us to deal with general natural images that contain complex structures and textures, without the need to 
rely on the existence of a database of images from the same class. This is achieved by a pyramid of fully convolutional light-weight GANs, each is responsible for capturing the distribution of patches at a different scale.  Once trained, \emph{SinGAN} can produce diverse high quality image samples (of arbitrary dimensions), which semantically resemble the training image, yet contain new object configurations and structures\footnote{Code available at: \url{https://github.com/tamarott/SinGAN}} (Fig.~\ref{fig:teaser}).

Modeling the internal distribution of patches within a single natural image has been long recognized as a powerful prior in many computer vision tasks \cite{zontak2011internal}. Classical examples include denoising \cite{zontak2013separating}, deblurring \cite{michaeli2014blind}, super resolution \cite{glasner2009super}, dehazing \cite{bahat2016blind,freedman2011image}, and image editing \cite{mechrez2018saliency,he2012statistics,cho2008patch, dekel2015revealing, tlusty2018modifying}. The most closley related work in this context is \cite{simakov2008summarizing}, where a bidirectional patch similarity measure is defined and optimized to guarantee that the patches of an image after manipulation are the same as the original ones.  Motivated by these works, here we show how SinGAN can be used within a simple unified learning framework to solve a variety of image manipulation tasks, including paint-to-image, editing, harmonization, super-resolution, and animation from a single  
image. In all these cases, our model produces high quality results that preserve the internal patch statistics of the training image (see Fig.~\ref{fig:image_manipulation} and our \href{http://webee.technion.ac.il/people/tomermic/SinGAN/SinGAN.htm}{project webpage}). All tasks are achieved with \emph{the same} generative network, without any additional information or further training beyond the original training image.

%% file: related.tex
\subsection{Related Work}

\begin{figure}[t]
	\centering
	\includegraphics[width=0.89\columnwidth]{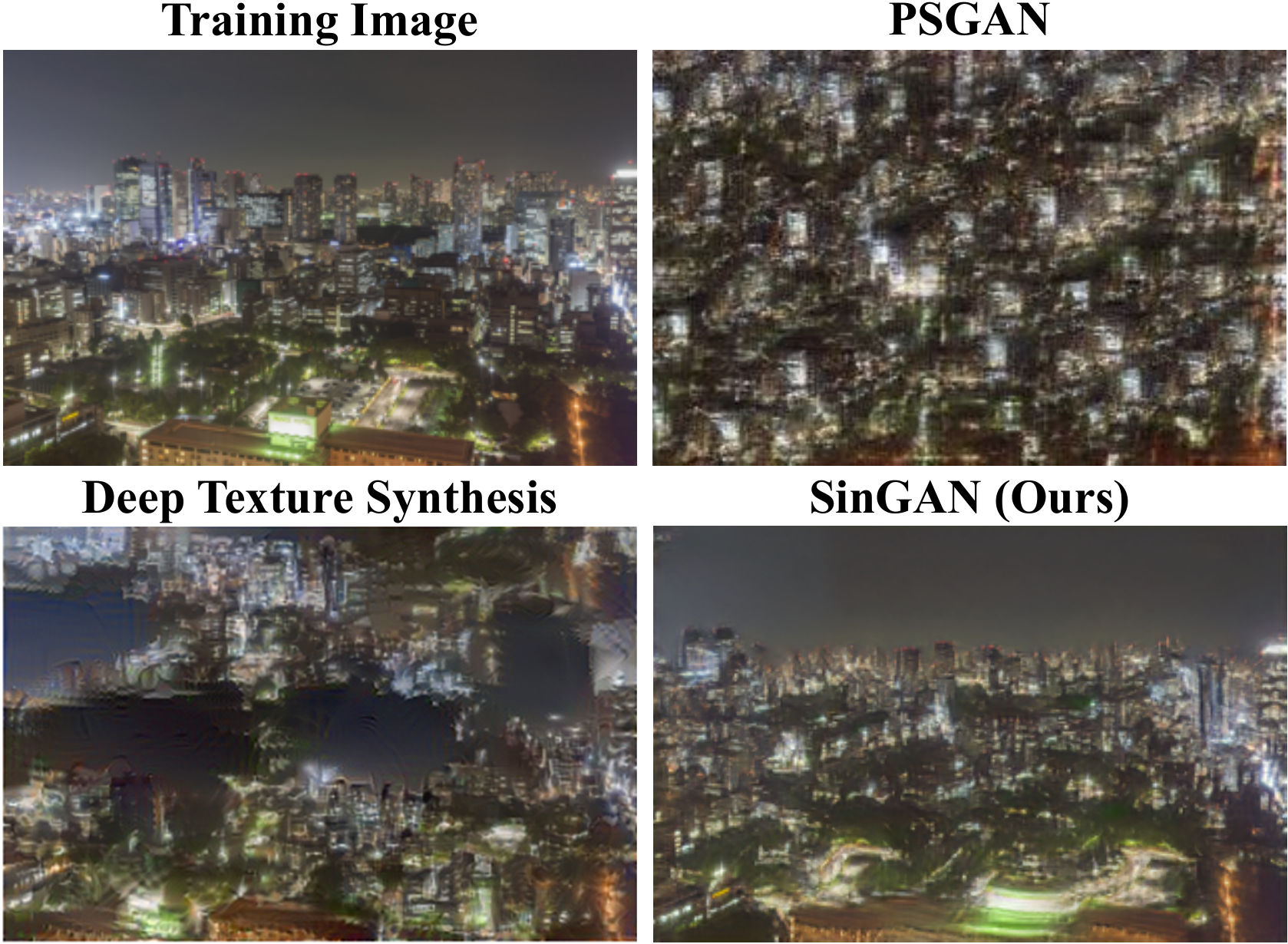}
	\caption{\textbf{SinGAN vs.~Single Image Texture Generation.} Single image models for texture generation~\cite{bergmann2017learning, gatys2015texture} are not designed to deal with natural images.  Our model can produce realistic image samples that consist of complex textures and non-reptititve global structures.}
	\label{fig:texture GAN}
	\vspace{-0.6cm}
\end{figure}

\paragraph{Single image deep models} Several recent works proposed to ``overfit'' a deep model to a single training example  \cite{ulyanov2017deep,zhou2018non,shocher2018zero, chan2018dance, asano2019surprising}. However, these methods are designed for specific tasks (\eg, super resolution~\cite{shocher2018zero}, texture expansion~\cite{zhou2018non}). Shocher~\etal~\cite{shocher2018internalArxiv,shocher2018internal} were the first to introduce an internal GAN based model for a single natural image, and illustrated it in the context of retargeting. However, their generation is conditioned on an input image (\ie, mapping images to images) and is not used to draw random samples. 
In contrast, our framework is 
purely generative (\ie maps noise to image samples), and thus suits many different image manipulation tasks.
\emph{Unconditional} single image GANs have been explored only in the context of texture generation \cite{bergmann2017learning,jetchev2016texture,li2016precomputed}. These models do not generate meaningful samples when trained on non-texture images (Fig.~\ref{fig:texture GAN}). Our method, on the other hand, is not restricted to texture and can handle general natural images (\eg, Fig.~\ref{fig:teaser}). 
 
\begin{figure*}[t]
\vspace{-0.5cm}
	\centering
	\includegraphics[width=0.85\textwidth]{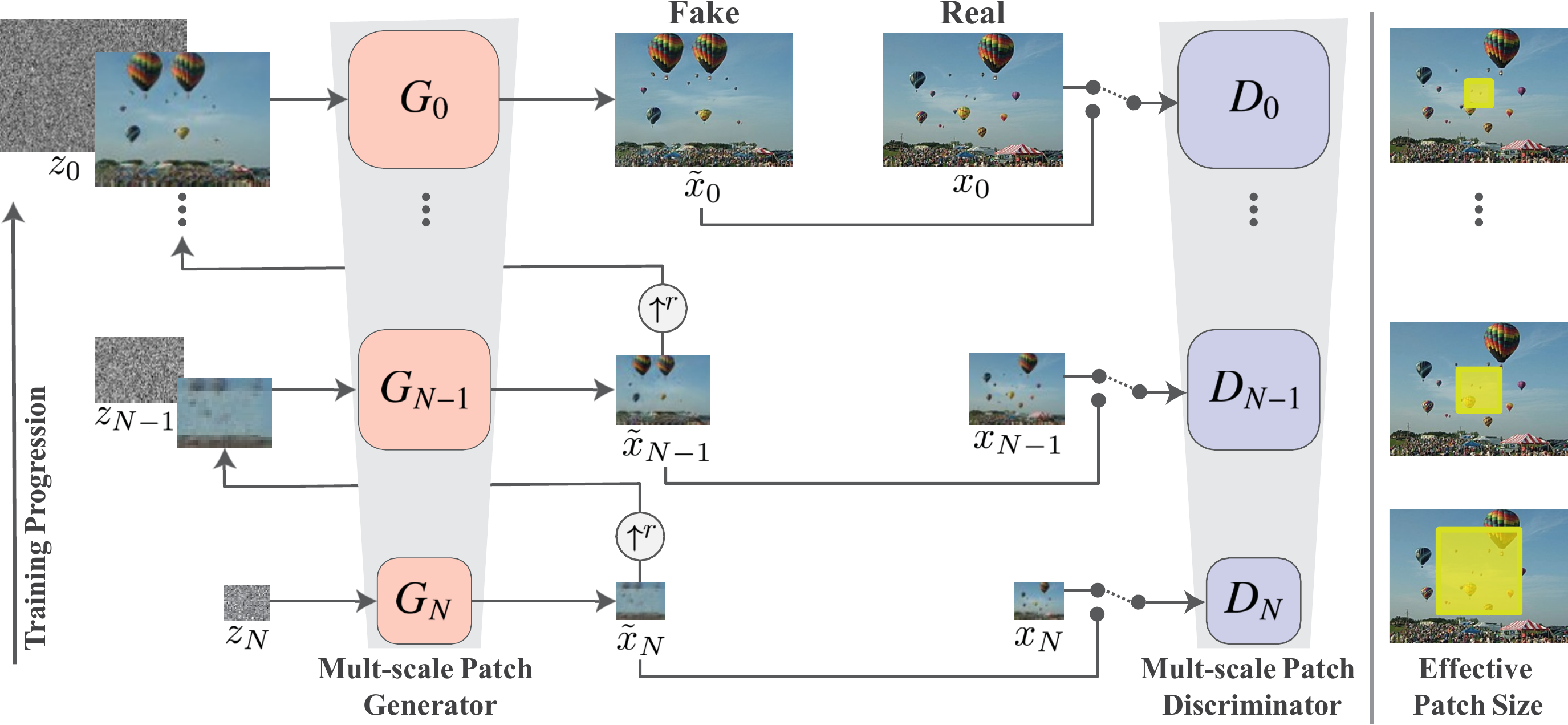}
	\caption{\textbf{SinGAN's multi-scale pipeline.} Our model consists of a pyramid of GANs, where both training and inference are done in a coarse-to-fine fashion. At each scale, $G_n$ learns to generate image samples in which all the overlapping patches cannot be distinguished from the patches in the down-sampled training image, $x_n$, by the discriminator $D_n$; the effective patch size decreases as we go up the pyramid (marked in yellow on the original image for illustration). The input to $G_n$ is a random noise image $z_n$, and the generated image from the previous scale $\tilde{x}_n$, upsampled to the current resolution (except for the coarsest level which is purely generative). The generation process at level $n$ involves all generators $\{G_N \ldots G_n\}$ and all noise maps $\{z_N,\ldots,z_n\}$ up to this level. See more details at Sec.~\ref{sec:method}.
	}
	\label{fig:architecture} \vspace{-0.4cm}
\end{figure*}

\begin{figure}[t]
	\centering
	\includegraphics[width=1\columnwidth]{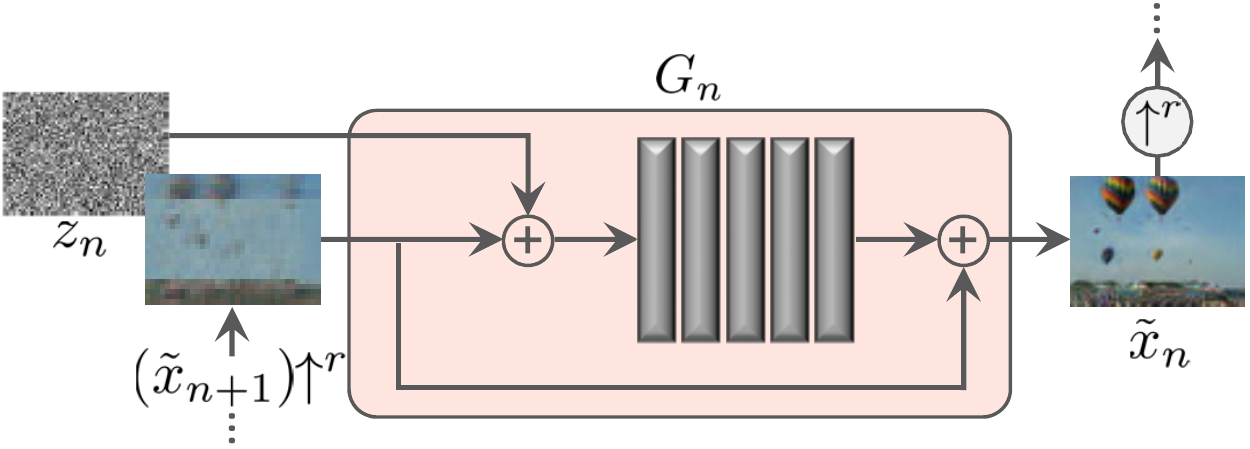}
	\caption{\textbf{Single scale generation.} At each scale $n$, the image from the previous scale, $\tilde{x}_{n+1}$, is upsampled and added to the input noise map, $z_n$. The result is fed into $5$ conv layers, whose output is a residual image that is added back to $(\tilde{x}_{n+1})\uparrow^r$. This is the output $\tilde{x}_n$ of $G_n$.}
	\label{fig:single_scale} \vspace{-0.4cm}
\end{figure} 
 
 %
 


\vspace{-0.4cm}

\paragraph{Generative models for image manipulation}
The power of adversarial learning has been demonstrated by recent GAN-based methods, in many different image manipulation tasks~\cite{zhu2016generative, dekel2018sparse, zhu2017unpaired, chen2018sketchygan, wang2016generative, yu2018generative, perarnau2016invertible, wang2016generative}. 
Examples include interactive image editing \cite{zhu2016generative, dekel2018sparse}, sketch2image~\cite{chen2018sketchygan, sangkloy2017scribbler}, and other image-to-image translation tasks~\cite{zhu2017unpaired, wang2017high, Xian_2018_CVPR}. However, all these methods are trained on class specific datasets, and here too, often condition the generation on another input signal. We are not interested in capturing common features among images of the same class, but rather consider a different source of training data -- all the overlapping patches at multiple scales of a single natural image.  We show that a powerful generative model can be learned from this data, and can be used in a number of image manipulation tasks.

%% file: method.tex
\section{Method}
\label{sec:method}
Our goal is to learn an \emph{unconditional} generative model that captures the internal statistics of a \emph{single} training image $x$. This task is conceptually similar to the conventional GAN setting, except that here the training samples are patches of a single image, rather than whole image samples from a database. 

We opt to go beyond texture generation, and to deal with more general natural images. This requires capturing the statistics of complex image structures at many different scales. For example, we want to capture global properties such as the arrangement and shape of large objects in the image (\eg sky at the top, ground at the bottom), as well as fine details and texture information. To achieve that, our generative framework, illustrated in Fig.~\ref{fig:architecture}, consists of a hierarchy of patch-GANs (Markovian discriminator) \cite{li2016precomputed,isola2017image}, where each is responsible for capturing the patch distribution at a different scale of $x$. The GANs have small receptive fields and limited capacity, preventing them from memorizing the single image. 
While similar multi-scale architectures have been explored in conventional GAN settings (\eg \cite{karras2017progressive,wang2017high,karras2018style,wang2017high,denton2015deep,huang2017stacked}), we are the first explore it for internal learning from a single image.


\subsection{Multi-scale architecture}
Our model consists of a pyramid of generators, $\{G_0, \ldots, G_N \}$, trained against an image pyramid of~ 
$x$: $\{x_0,\ldots,x_N\}$, where $x_n$ is a downsampled version of~$x$ by a factor $r^n$, for some $r>1$. 
Each generator $G_n$ is responsible of producing realistic image samples w.r.t.\@ the patch distribution in the corresponding image $x_n$. This is achieved through adversarial training, where $G_n$ learns to fool an associated 
discriminator $D_n$, which attempts to distinguish patches in the generated samples from patches in~$x_n$.

The generation of an image sample starts at the coarsest scale and sequentially passes through all generators up to the finest scale, with noise injected at every scale. All the generators and discriminators have the same receptive field and thus capture structures of decreasing size as we go up the generation process. At the coarsest scale, the generation is purely generative, \ie $G_N$ maps spatial white Gaussian noise $z_N$ to an image sample $\tilde{x}_N$, \vspace{-0.1cm}
\begin{align}
\vspace{-0.3cm}
\label{eq:GenChain}
\tilde{x}_N= G_N(z_N).
\end{align}
The effective receptive field at this level is typically $\sim1/2$ of the image's height, hence $G_N$ generates the general layout of the image and the objects' global structure.  
Each of the generators $G_n$ at finer scales ($n<N$) adds details that were not generated by the previous scales.  
Thus, in addition to spatial noise $z_n$, each generator $G_n$ accepts an upsampled version of the image from the coarser scale, \ie,  
\begin{align}
\tilde{x}_{n} = G_n\left(z_n,\,(\tilde{x}_{n+1})\uparrow^r\right),\qquad n<N.
\end{align}

All the generators have a similar architecture, as depicted in Fig.~\ref{fig:single_scale}. Specifically, the noise $z_n$ is added to the image $(\tilde{x}_{n+1})\uparrow^r$, prior to being fed into a sequence of convolutional layers. This ensures that the GAN does not disregard the noise, as often happens in conditional schemes involving randomness \cite{zhu2017unpaired,mathieu2015deep,zhu2017toward}. The role of the convonlutional layers is to generate the missing details in~$(\tilde{x}_{n+1})\uparrow^r$ (residual learning \cite{he2016deep,zhang2017beyond}). Namely, $G_n$ performs the operation
\begin{equation}\label{eq:}
    \tilde{x}_{n} = (\tilde{x}_{n+1})\uparrow^r  + \,\, \psi_n\left(z_n + (\tilde{x}_{n+1})\uparrow^r\right),
\end{equation}
where $\psi_n$ is a fully convolutional net 
with 5 conv-blocks of the form Conv($3\times 3$)-BatchNorm-LeakyReLU \cite{ioffe2015batch}. 
We start with $32$ kernels per block at the coarsest scale and increase this number by a factor of $2$ every $4$ scales. Because the generators are fully convolutional, we can generate images of arbitrary size and aspect ratio at test time (by changing the dimensions of the noise maps).

\begin{figure*}
	\centering
	\includegraphics[width=1\textwidth]{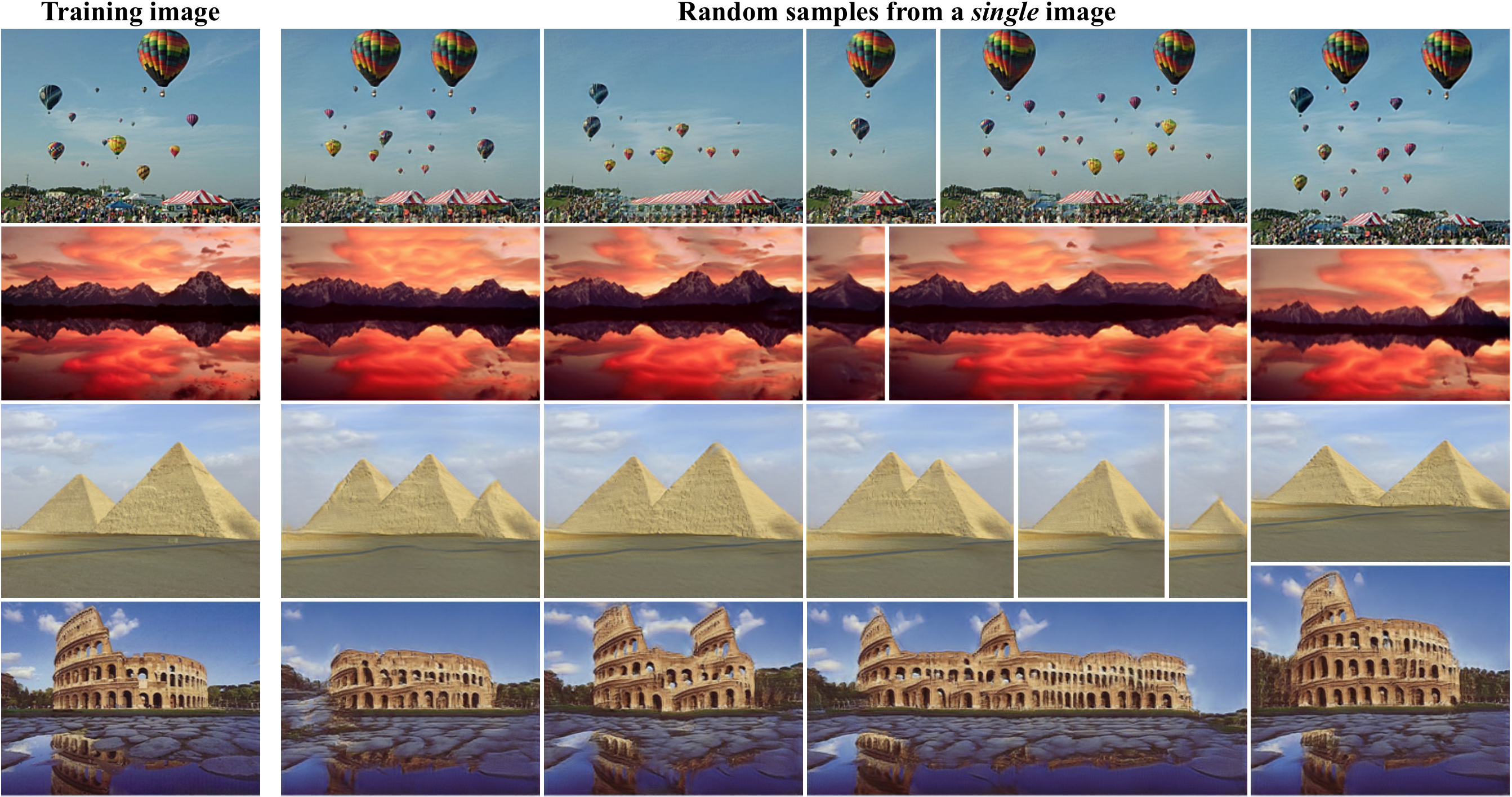}
	\caption{\textbf{Random image samples.} After training SinGAN on a single image, our model can generate realistic random image samples that depict new structures and object configurations, yet preserve the patch distribution of the training image. Because our model is fully convolutional, the generated images may have arbitrary sizes and aspect ratios. Note that our goal is not image retargeting -- our image samples are random and optimized to maintain the patch statistics, rather than preserving
salient objects. See SM for more results and qualitative comparison to image retargeting methods.}
	\label{fig:samples} \afterfig
\end{figure*}

\subsection{Training}
We train our multi-scale architecture sequentially, from the coarsest scale to the finest one. Once each GAN is trained, it is 
kept fixed. 
Our training loss for the $n$th GAN is comprised of an adversarial term and a reconstruction term,
\begin{equation}
\label{eq:total Loss}
\min_{G_n}\,\,\max_{D_n}\,\, \mathcal{L}_{\text{adv}}(G_n,D_n)+\alpha\mathcal{L}_{\text{rec}}(G_n).
\end{equation}
The adversarial loss $\mathcal{L}_{\text{adv}}$ penalizes for the distance between the distribution of patches in $x_n$ and the distribution of patches in generated samples $\tilde{x}_n$. 
The reconstruction loss $\mathcal{L}_{\text{rec}}$ insures the existence of a specific set of noise maps that can produce $x_n$, an important feature for image manipulation (Sec.~\ref{sec:applications}). 
We next describe $\mathcal{L}_{\text{adv}},\mathcal{L}_{\text{rec}}$ in detail. See Supplementary Materials (SM) for optimization details.

\begin{figure*}
	\centering
	\includegraphics[width=1\textwidth]{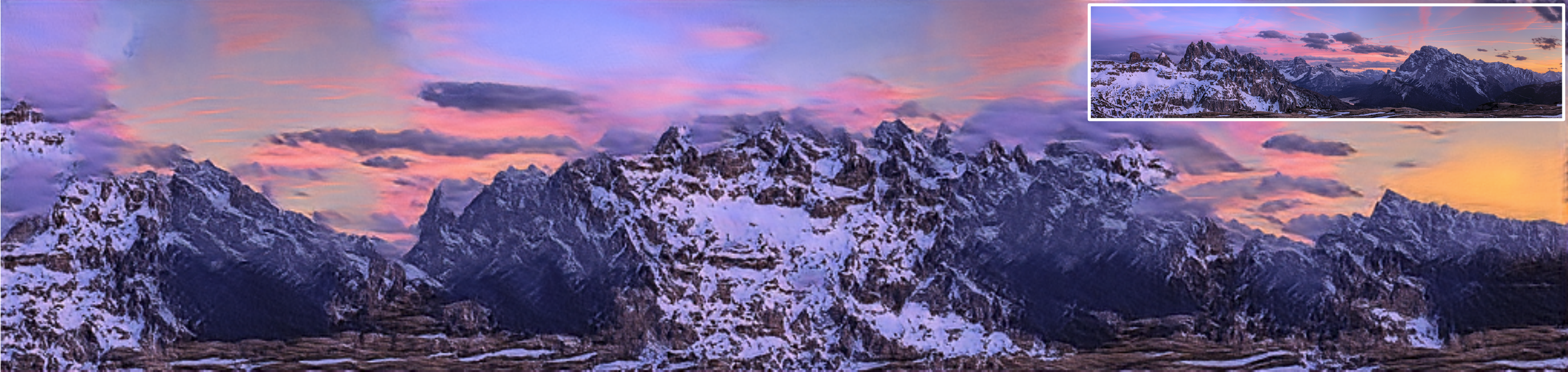}
	\caption{\textbf{High resolution image generation.} A random sample produced by our model, trained on the $243\times1024$ image (upper right corner); new global structures as well as fine details are realistically generated. See 4Mpix examples in SM.} 
	\label{fig:high-res} \afterfig
\end{figure*}

\vspace{-0.2cm}
\paragraph{Adversarial loss}  Each of the generators $G_n$ is coupled with a Markovian discriminator $D_n$ that classifies each of the overlapping patches of its input as real or fake \cite{li2016precomputed,isola2017image}. We use the WGAN-GP loss \cite{gulrajani2017improved}, which we found to increase training stability, where the final discrimination score is the average over the patch discrimination map. 
As opposed to single-image GANs for textures (\eg, \cite{li2016precomputed,jetchev2016texture,bergmann2017learning}), here we define the loss over the whole image rather than over random crops (a batch of size~$1$). This allows the net to learn boundary conditions (see SM), which is an important feature in our setting. The architecture of $D_n$ is the same as the net $\psi_n$ within $G_n$
, so that its patch size (the net's receptive field) is $11\times11$.

\vspace{-0.2cm}
\paragraph{Reconstruction loss}
We want to ensure that there exists a specific set of input noise maps, which generates the original image $x$. We specifically choose $\{z_N^{\text{rec}},z_{N-1}^{\text{rec}}\ldots,z_0^{\text{rec}}\}=\{z^*,0,\ldots,0\}$, where $z^*$ is some fixed noise map (drawn once and kept fixed during training). Denote by $\tilde{x}_n^{\text{rec}}$ the generated image at the $n$th scale when using these noise maps. Then for $n<N$, 
\begin{equation}
\label{eq:rec}
 \mathcal{L}_{\text{rec}}=\|G_n(0, (\tilde{x}_{n+1}^{\text{rec}})\uparrow^r)- x_n\|^2,
\end{equation}
and for $n=N$, we use $\mathcal{L}_{\text{rec}}= \|G_N(z^*)- x_N\|^2$.

The reconstructed image $\tilde{x}_n^{\text{rec}}$ has another role during training, which is to determine the standard deviation $\sigma_n$ of the noise $z_n$ in each scale. Specifically, we take $\sigma_n$ to be proportional 
to the root mean squared error (RMSE) between $(\tilde{x}_{n+1}^{\text{rec}})\uparrow^r$ and $x_n$, which gives an indication of the amount of details that need to be added at that scale.

%% file: evaluation.tex
\section{Results}

We tested our method both qualitatively and quantitatively on a variety of images spanning a large range of scenes including urban and nature scenery as well as artistic and texture images. The images that we used are taken from the Berkeley Segmentation Database (BSD) \cite{martin2001database}, Places \cite{zhou2014learning} and the Web. We always set the minimal dimension at the coarsest scale to $25$px, and choose the number of scales $N$ s.t.\@ the scaling factor $r$ is as close as possible to $4/3$. For all the results, (unless mentioned otherwise), we resized the training image to maximal dimension $250$px.

\begin{figure}
	\centering
	\includegraphics[width=1\columnwidth]{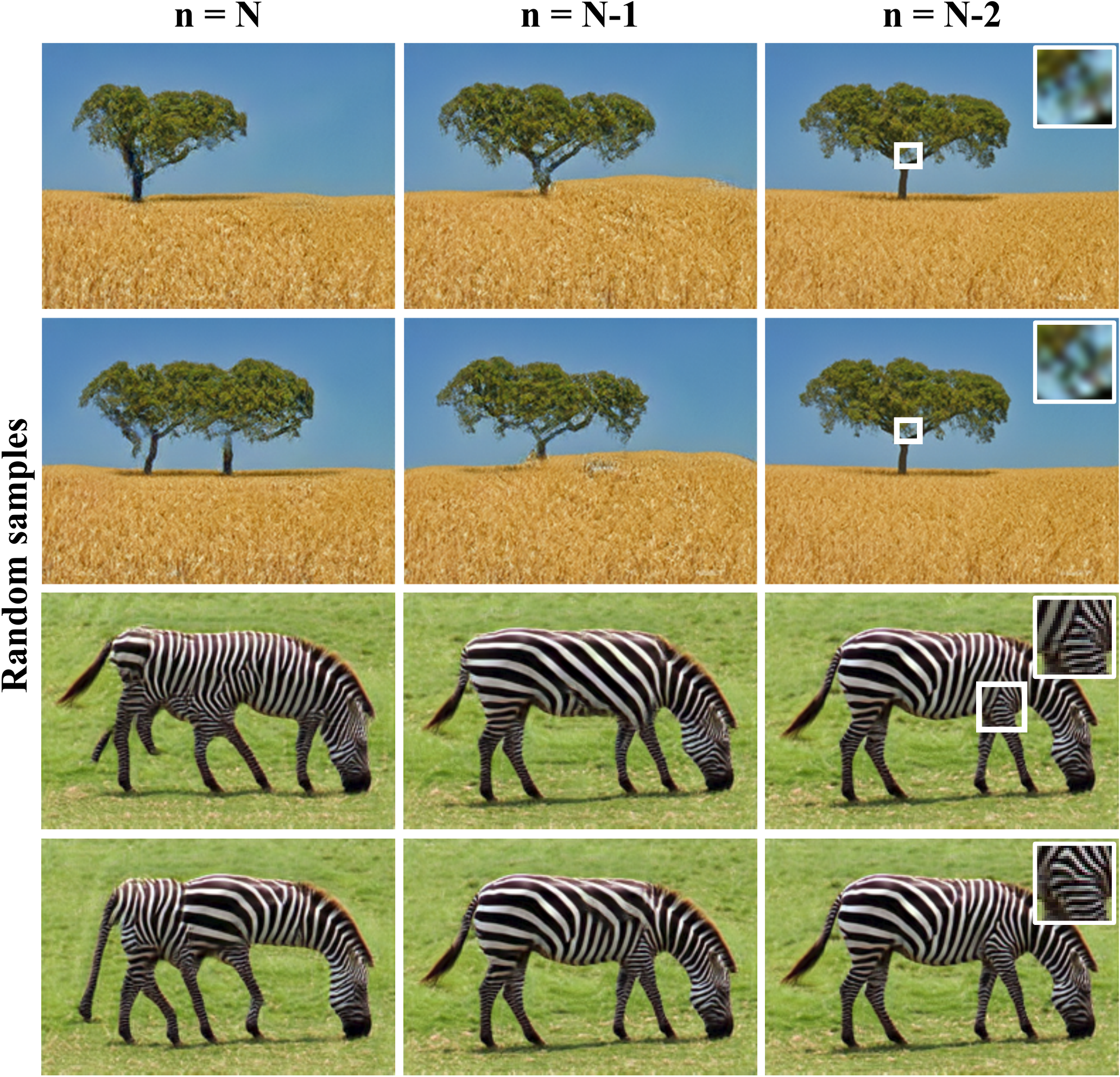}
	\caption{\textbf{Generation from different scales (at inference).} We show the effect of starting our hierarchical generation from a given level $n$. For our full generation scheme ($n=N$), the input at the coarsest level is random noise. For generation from a finer scale $n$, we plug in the downsampled original image, $x_n$, as input to that scale. This allows us to control the scale of the generated structures, \eg, we can preserve the shape and pose of the Zebra and only change its stripe texture by starting the generation from $n=N-1$.
	}
	\label{fig:variability}\afterfig
\end{figure}

Qualitative examples of our generated random image samples are shown in Fig.~\ref{fig:teaser}, Fig.~\ref{fig:samples}, 
and many more examples are included in the SM. For each example, we show a number of random samples with the same aspect ratio as the original image, and with decreased and expanded dimensions in each axis. As can be seen, in all these cases, the generated samples depict new realistic structures and configuration of objects, while preserving the visual content of the training image. Our model successfully preservers global structure of objects, \eg mountains~(Fig.~\ref{fig:teaser}), air balloons or pyramids~(Fig.~\ref{fig:samples}), as well as fine texture information. Because the network has a limited receptive field (smaller than the entire image), it can generate new combinations of patches that do not exist in the training image 
Furthermore, we observe that in many cases reflections and shadows are realistically synthesized, as can be seen in Fig.~\ref{fig:samples} and Fig.~\ref{fig:teaser} (and the first example of Fig.~\ref{fig:variability}). 
Note that SinGAN's architecture is resolution agnostic and can thus be used on high resolution images, as illustrated in Fig.~\ref{fig:high-res} (see 4Mpix results in the SM). Here as well, structures at all scales are nicely generated, from the global arrangement of sky, clouds and mountains, to the fine textures of the snow.


\vspace{-0.3cm}
\paragraph{Effect of scales at test time}
Our multi-scale architecture allows control over the amount of variability between samples, by choosing the scale from which to start the generation at test time. To start at scale $n$, we fix the noise maps up to this scale to be $\{z_N^{\text{rec}},\ldots,z^{\text{rec}}_{n+1}\}$, and use random draws only for $\{z_n,\ldots,z_0\}$. The effect is illustrated in Fig.~\ref{fig:variability}. As can be seen, starting the generation at the coarsest scale ($n=N$), results in large variability in the global structure. In certain cases with a large salient object, like the Zebra image, this may lead to unrealistic samples. However, starting the generation from finer scales, enables to keep the global structure intact, while altering only finer image features (\eg the Zebra's stripes). See SM for more examples.

\vspace{-0.3cm}
\paragraph{Effect of scales during training}

\begin{figure}[t]
	\centering
	\includegraphics[width=1\columnwidth]{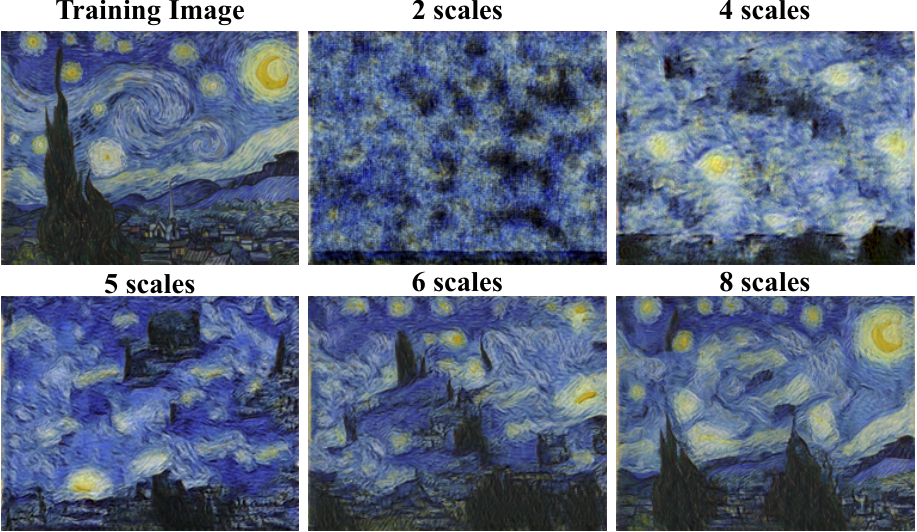}
	\caption{\textbf{The effect of training with a different number of scales.} 
	The number of scales in SinGAN's 
	architecture strongly influences the results. A model with a small number of scales only captures textures. As the number of scales increases, SinGAN manages to capture larger structures as well as the global arrangement of objects in the scene.}
	\label{fig:multi_scale} \afterfig
\end{figure}

Figure~\ref{fig:multi_scale} shows the effect of training with fewer scales. With a small number of scales, the effective receptive field at the coarsest level is smaller, allowing to capture only fine textures. As the number of scales increases, structures of larger support emerge, and the global object arrangement is better preserved.

\subsection{Quantitative Evaluation}
To quantify the realism of our generated images and how well they capture the internal 
statistics of the training image, we use two metrics: (i)  Amazon Mechanical Turk (AMT) ``Real/Fake'' user study, and (ii) a new single-image version of the Fr\'echet Inception Distance \cite{heusel2017gans}.
  
\vspace{-0.4cm}
\paragraph{AMT perceptual study} 
We followed the protocol of \cite{isola2017image, zhang2016colorful} and performed perceptual experiments in 2 settings. (i) Paired (real vs.~fake): Workers were presented with a sequence of 50 trials, in each of which a fake image (generated by SinGAN) was presented against its real training image for 1 second. Workers were asked to pick the fake image. (ii) Unpaired (either real or fake): Workers were presented with a \emph{single} image for 1 second, and were asked if it was fake. In total, 50 real images and a disjoint set of 50 fake images were presented in random order to each worker. 

We repeated these two protocols for two types of generation processes: Starting the generation from the coarsest ($N$th) scale, and 
from scale $N-1$ (as in Fig.~\ref{fig:variability}). 
This way, we assess the realism of our results in two different variability levels. To quantify the diversity of the generated images, for each training example we calculated the standard deviation (std) of the intensity values of each pixel over 100 generated images, averaged it over all pixels, and normalized by the std of the intensity values 
of the training image.

The real images were randomly picked from the ``places'' database \cite{zhou2014learning} from the subcategories Mountains, Hills, Desert, Sky. In each of the 4 tests, we had 50 different participants. In all tests, the first 10 trials were a tutorial including a feedback. The results are reported in Table \ref{tab:AMT}.

\begin{table}[t]
    \centering
    \begin{tabular}{|c|c|c|c|}
    \hline
    1st Scale & Diversity & Survey & Confusion \\
    \hline
    $N$ & 0.5 & \multirow{2}{3.3em}{paired\\ unpaired} & $21.45\% \pm 1.5\%$ \\
    & & &$42.9\% \pm 0.9\% $ \\
    \hline
    $N-1$ & 0.35 & \multirow{2}{3.3em}{paired\\ unpaired}& $30.45\% \pm 1.5\% $ \\
    & & &$47.04\% \pm 0.8\% $ \\
    \hline
    \end{tabular}
    \caption{{\bf ``Real/Fake'' AMT test.} We report confusion rates for two generation processes: Starting from the coarsest scale $N$ (producing samples with large diversity), and starting from the second coarsest scale $N\!\!-\!\!1$ (preserving the global structure of the original image). In each case, we performed both a paired study (real-vs.-fake image pairs are shown), and an unpaired one (either fake or real image is shown). The variance was estimated by bootstrap~\cite{efron1992bootstrap}.}
    \label{tab:AMT}\afterfig
\end{table}
As expected, the confusion rates are consistently larger in the unpaired case, where there is no reference for comparison. In addition, it is clear that the confusion rate decreases with the diversity of the generated images. However, even when large structures are changed, our generated images were hard to distinguish from the real 
images (a score of 50\% would mean perfect confusion between real and fake). The full set of test images are included in the SM. 


\begin{figure*}[t!]
	\centering
	\includegraphics[width=1\textwidth]{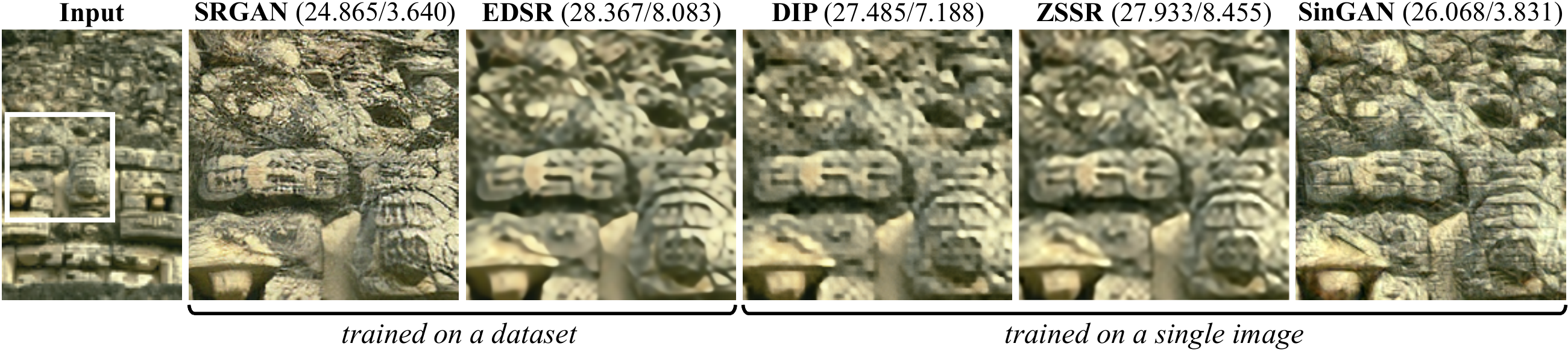}
	\caption{\textbf{Super-Resolution.} When SinGAN is trained on a low resolution image, we are able to super resolve. This is done by iteratively upsampling the image and feeding it to SinGAN's finest scale generator. As can be seen, SinGAN's visual quality is better than the SOTA internal methods ZSSR \cite{shocher2018zero} and DIP \cite{ulyanov2017deep}. It is also better than EDSR \cite{lim2017enhanced} and comparable to SRGAN \cite{ledig2017photo}, external methods trained on large collections. Corresponding PSNR and NIQE \cite{mittal2013making} are shown in parentheses.}
	\label{fig:sr} \vspace{-0.20cm}
\end{figure*}
\begin{figure*}[t!]
	\centering
	\includegraphics[width=1\textwidth]{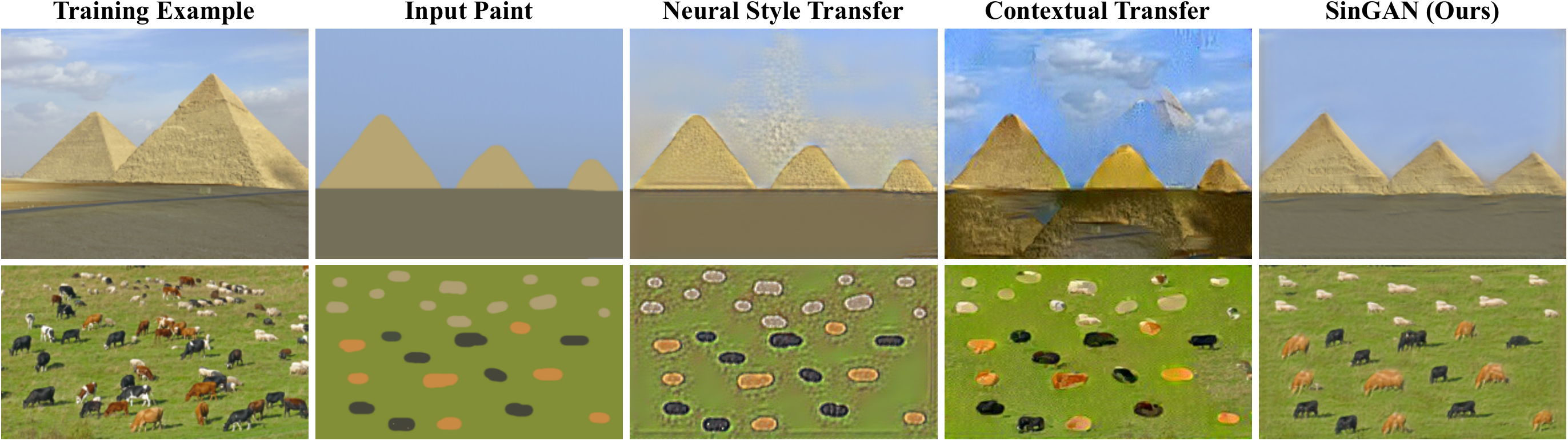}
	\caption{\textbf{Paint-to-Image.} We train SinGAN on a target image and inject a downsampled version of the paint into one of the coarse levels at test time. Our generated images preserve the layout and general structure of the clipart while generating realistic texture and fine details that match the training image. Well-known style transfer methods~\cite{gatys2016image,mechrez2018contextual} fail in this task.}\afterfig
	\label{fig:Paint}
\end{figure*}

\vspace{-0.35cm}
\paragraph{Single Image Fr\'echet Inception Distance}
\noindent We next quantify how well SinGAN captures the internal statistics of $x$. A common metric for GAN evaluation is the Fr\'echet Inception Distance (FID) \cite{heusel2017gans}, which measures the deviation between the distribution of deep features of generated images and that of real images. In our setting, however, we only have a single real image, and are rather interested in its \emph{internal} patch statistics. We thus propose the Single Image FID (SIFID) metric. 
Instead of using the activation vector after the last pooling layer in the Inception Network \cite{szegedy2015going} (a single vector per image), we use the internal distribution of deep features at the output of the convolutional layer just before the second pooling layer (one vector per location in the map). 
Our SIFID is the FID between the statistics of those features in the real image and in the generated sample. 

As can be seen in Table~\ref{tab:FID}, the average SIFID is lower for generation from scale $N\!-\!1$ than for generation from scale $N$, which aligns with the user study results. 
We also report the correlation between the SIFID scores and the confusion rates for the fake images. Note that there is a significant (anti) correlation between the two, implying that a small SIFID is typically a good indicator for a large confusion rate. The correlation is stronger for the paired tests, since SIFID is a paired measure (it operates on the pair $x_n,\tilde{x}_n$).


%% file: applications.tex
{\small 
\begin{table}[t]
    \centering
    \begin{tabular}{|c|c|c|c|}
    \hline
         1st Scale & SIFID & Survey &  SIFID/AMT Correlation\\
         \hline
         $N$ & 0.09 & \multirow{2}{3.3em}{paired\\ unpaired} & $-0.55$ \\
          & & & $-0.22$ \\
         \hline
         $N-1$ & 0.05 & \multirow{2}{3.3em}{paired \\ unpaired}  & $-0.56$ \\
         & & & $-0.34$ \\
         \hline
    \end{tabular}
    \caption{\textbf{Single Image FID (SIFID).} We adapt the FID metric to a 
    single image 
    and report the average score for 50 images, for full generation (first row), and starting from the second coarsest scale (second row).
    Correlation with AMT results 
    shows SIFID highly agrees with human ranking.
    }
    \label{tab:FID}\afterfig \vspace{-0.2cm}
\end{table}}

\vspace{-0.1cm}
\section{Applications}
\label{sec:applications}


We explore the use of SinGAN for a number of image manipulation tasks. To do so, we use our model \emph{after training}, with no architectural changes or further tuning and follow the same 
approach for all applications. The idea is to utilize the fact that at inference, SinGAN can only produce images with the same patch distribution as the training image. Thus, 
manipulation can be done by injecting (a possibly downsampled version of) an image into the generation pyramid at some scale $n<N$, and feed forwarding 
it through the generators so as to match its patch distribution to that of the training image. Different injection scales lead to different effects. 
We consider the following applications (see SM for more results and the injection scale effect). 

\vspace{-0.4cm}
\paragraph{Super-Resolution}
\emph{Increase the resolution of an input image by a factor $s$.} 
We train our model on the low-resolution (LR) image, with a reconstruction loss weight of $\alpha=100$ and a pyramid scale factor of $r=\sqrt[k]{s}$ for some $k\in\mathbb{N}$. Since small structures tend to recur across scales of natural scenes \cite{glasner2009super}, at test time we upsample the LR image by a factor of $r$ and inject it (together with noise) to the last generator, $G_0$. We repeat this $k$ times to obtain the final high-res output. 
An example result is shown in Fig.~\ref{fig:sr}. As can be seen, the visual quality of our reconstruction exceeds that of state-of-the-art \emph{internal} methods \cite{ulyanov2017deep,shocher2018zero} as well as of \emph{external} methods that aim for PSNR maximization \cite{lim2017enhanced}. Interestingly, it is comparable to the externally trained SRGAN method \cite{ledig2017photo}, despite having been exposed to only a single image. Following~\cite{blau20182018}, we compare these 5 methods in Table~\ref{tab:SR} on the BSD100 dataset \cite{martin2001database} in terms of distortion (RMSE) and perceptual quality (NIQE \cite{mittal2013making}), which are two fundamentally conflicting requirements \cite{blau2018perception}. As can be seen, SinGAN excels in perceptual quality; its NIQE score is only slightly inferior to SRGAN, and its RMSE is slightly better.  


\begin{table}[t]
    \centering
    \begin{tabular}{|c|c|c|c|c|c|}
    \hline
    & \multicolumn{2}{c|}{External methods} & \multicolumn{3}{c|}{Internal methods} \\
    \hline
    & SRGAN & EDSR & DIP & ZSSR & SinGAN \\ 
    \hline
    RMSE & 16.34 & 12.29 & 13.82 & 13.08 & 
    16.22 \\
    \hline
    NIQE & 3.41 & 6.50 & 6.35 & 7.13 & 
    3.71 \\
    \hline
    \end{tabular}\vspace{-0.1cm}
    \caption{\textbf{Super-Resolution evaluation}. 
    Following \cite{blau2018perception}, we report distortion (RMSE) and perceptual quality (NIQE \cite{mittal2013making}, lower is better) on BSD100 \cite{martin2001database}. As can be seen, SinGAN's performance is similar to that of SRGAN \cite{ledig2017photo}.}
    \label{tab:SR}\afterfig 
\end{table}
\vspace{-0.4cm}

\paragraph{Paint-to-Image} \emph{Transfer a clipart into a photo-realistic image.} This is done by downsampling the clipart image and feeding it into one of the coarse scales (\eg $N\!-\!1$ or $N\!-\!2$). As can be seen in Figs.~\ref{fig:image_manipulation} and \ref{fig:Paint}, the global structure of the painting is preserved, while texture and high frequency information matching the original image are realistically generated. Our method outperforms style transfer methods \cite{mechrez2018contextual, gatys2016image} in terms of visual quality (Fig.~\ref{fig:Paint}).
\vspace{-0.4cm}

\begin{figure}[t!]
	\centering
	\includegraphics[width=1\columnwidth]{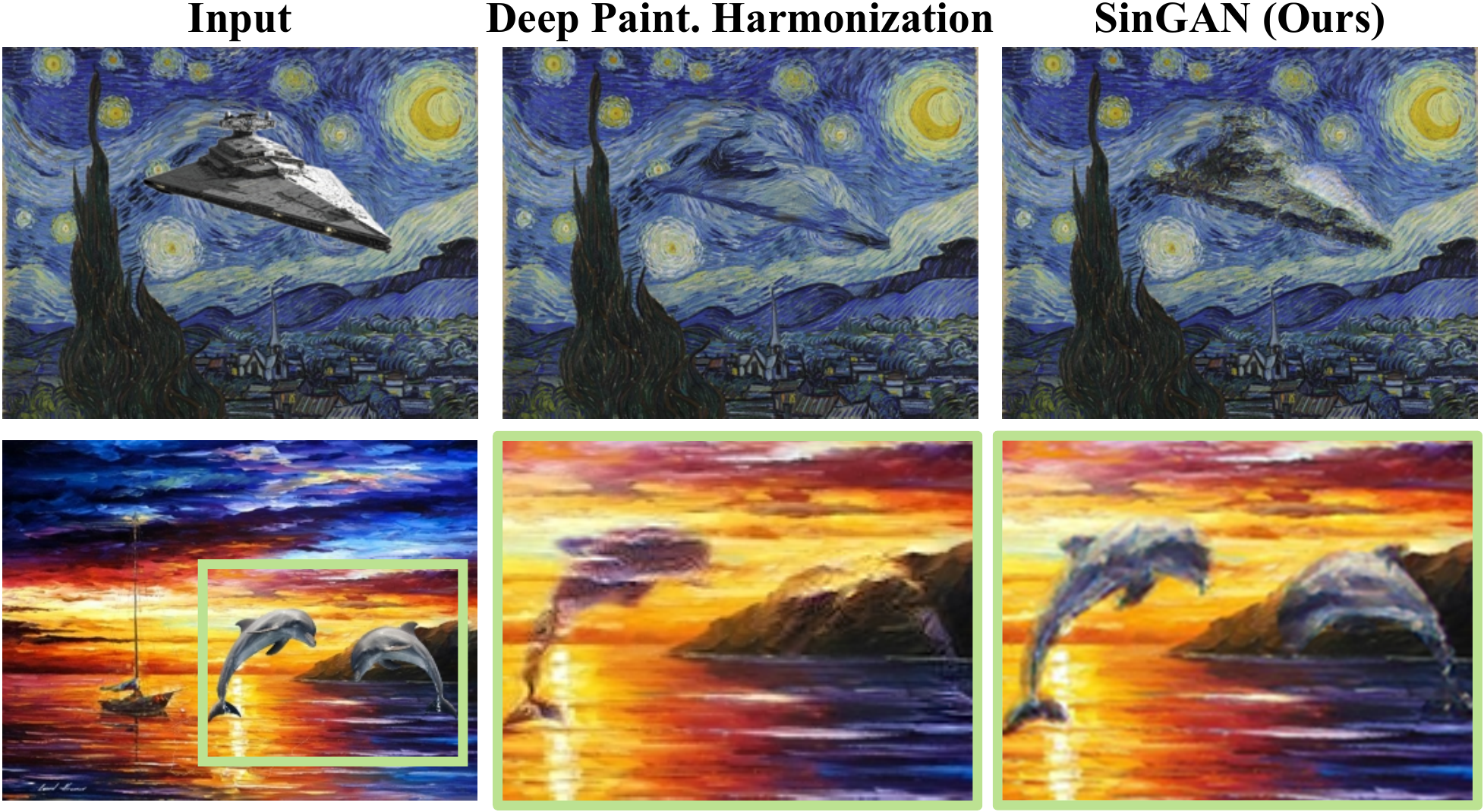}
	\caption{\textbf{Harmonization.} Our model is able to preserve the structure of the pasted object
	, while adjusting its appearance and texture. The dedicated harmonization method~\cite{luan2018deep} overly blends the object with the background.
	}\afterfig
	\label{fig:harmonization}
\end{figure}

\paragraph{Harmonization} \emph{Realistically blend a pasted object with a background image.} 
We train SinGAN on the background image, and inject a downsampled version of the naively pasted composite at test time. Here we combine the generated image with the original background. As can be seen in Fig.~\ref{fig:image_manipulation} and Fig.~\ref{fig:harmonization}, our model tailors the pasted object's texture to match the background, and often preserves its structure better than \cite{luan2018deep}.  
Scales 2,3,4 typically lead to good balance between preserving the object's structure and transferring the background's texture. 


\vspace{-0.4cm}

\begin{figure}[t!]
    \centering
	\includegraphics[width=1\columnwidth]{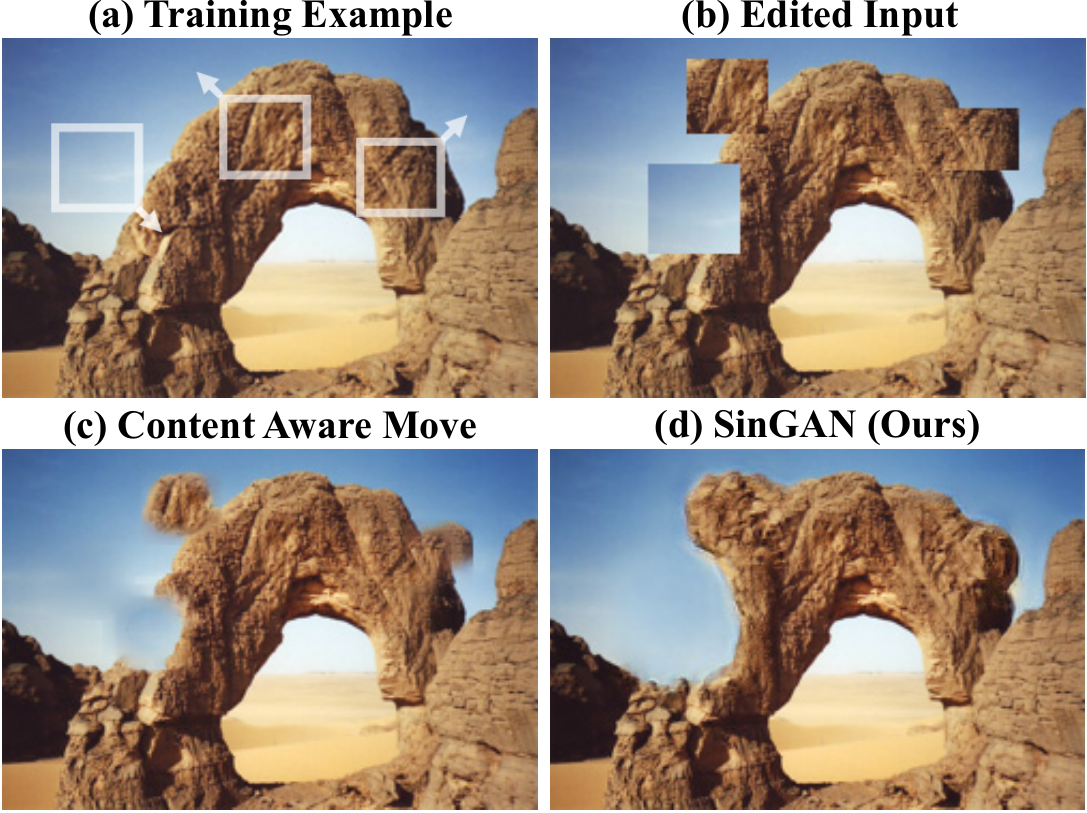}
	\caption{\textbf{Editing.} 
	We copy and paste a few patches from the original image (a), and input a downsampled version of the edited image (b) to an intermediate level of our model (pretrained on (a)). In the generated image (d), these local edits are translated into coherent and photo-realistic structures. (c)~comparison to Photoshop content aware move.}\afterfig
	\label{fig:Objects}
\end{figure}

\paragraph{Editing} \emph{Produce a seamless composite in which image regions have been copied and pasted in other locations.} 
Here, again, we inject a downsampled version of the composite into one of the coarse scales. We then combine SinGAN's output at the edited regions, with the original image. As shown in Fig.~\ref{fig:image_manipulation} and Fig.~\ref{fig:Objects}, SinGAN re-generates fine textures and seamlessly stitches the pasted parts, producing nicer results than Photoshop's Content-Aware-Move.

\vspace{-0.4cm}


\paragraph{Single Image Animation} 
\emph{Create a short video clip with realistic object motion, from a single input image.} Natural images often contain repetitions, which reveal different ``snapshots'' in time of the same dynamic object \cite{xu2008animating} (\eg an image of a flock of birds reveals all wing postures of a single bird). Using SinGAN, we can  travel along the manifold of all appearances of the object in the image, thus synthesizing motion from a single image. 
We found that for many types of images, a realistic effect is achieved by a random walk in $z$-space, starting with $z^{\text{rec}}$ for the first frame at all generation scales. Results are available on \url{https://youtu.be/xk8bWLZk4DU}. 


%% file: conclusions.tex
\section{Conclusion}
We introduced SinGAN, a new unconditional generative scheme that is learned from a single natural image. We demonstrated its ability to go beyond textures and to generate diverse realistic samples for natural complex images. Internal learning is inherently limited in terms of \emph{semantic} diversity compared to externally trained generation methods. For example, if the training image contains a single dog, our model will not generate samples of different dog breeds. Nevertheless, as demonstrated by our experiments, SinGAN can provide a very powerful tool for a wide range of image manipulation tasks.


